\pdfoutput=1
\documentclass[11pt]{article}

\usepackage[final]{coling}

\usepackage{times}
\usepackage{latexsym}
\usepackage[T1]{fontenc}
\usepackage[utf8]{inputenc}
\usepackage{microtype}
\usepackage{inconsolata}
\usepackage{graphicx}
\usepackage{todonotes}
\usepackage{subcaption}
\usepackage{multirow}

\usepackage{newfloat}
\usepackage{listings}

\DeclareFloatingEnvironment[%
    listname={List of Prompts},
    name=Prompt,
    placement=htb]{prompt}
\title{Searching for Structure: Investigating Emergent Communication with Large Language Models}
\author{
 \textbf{Tom Kouwenhoven\textsuperscript{1}},
 \textbf{Max Peeperkorn\textsuperscript{2}},
 and \textbf{Tessa Verhoef\textsuperscript{1}},
\\
 \textsuperscript{1}Leiden Institute of Advanced Computer Science, Leiden University, Netherlands,\\
 \textsuperscript{2}School of Computing, University of Kent, United Kingdom,
\\
 \small{
   \texttt{\{t.kouwenhoven, t.verhoef\}@liacs.leidenuniv.nl}
 }
}

\begin{document}
\maketitle
\begin{abstract}
Human languages have evolved to be structured through repeated language learning and use. These processes introduce biases that operate during language acquisition and shape linguistic systems toward communicative efficiency. In this paper, we investigate whether the same happens if artificial languages are optimised for implicit biases of Large Language Models (LLMs). To this end, we simulate a classical referential game in which LLMs learn and use artificial languages. Our results show that initially unstructured holistic languages are indeed shaped to have some structural properties that allow two LLM agents to communicate successfully. Similar to observations in human experiments, generational transmission increases the learnability of languages, but can at the same time result in non-humanlike degenerate vocabularies. Taken together, this work extends experimental findings, shows that LLMs can be used as tools in simulations of language evolution, and opens possibilities for future human-machine experiments in this field.
\end{abstract}

\section{Introduction}
Vocabularies of signals enable us to communicate about meanings, but to express an arbitrary number of meanings, vocabularies would require an equally large set of words as there are meanings, and learning such holistic vocabularies is cognitively challenging. Human languages therefore typically show some form of compositional structure, where meaningful signal-meaning mappings can be composed such that the combination of individual meaningful signals can express more than the meaning of the individual components alone \cite{hockett1960origin}. An important finding in the field of language evolution is that such structural properties can emerge as a result of individual learning biases and pressures that continuously shape the languages on a longer timescale, often eventually resulting in languages that are easier to learn and exhibit some degree of structure \cite{Smith2022HowStructure}. 

The processes involved in the evolution of language have been investigated abundantly with experiments and simulations. The latter typically use hard-coded agents with inductive biases \cite{deBoer2006computer}, Bayesian learners \cite[e.g.][]{griffits2007bayesianagents, culbertson2012bayesian, Kirby2015CompressionStructure}, or reinforcement learning agents \cite{lazaridou2020emergent} to investigate the evolution of structured languages. In contrast, we investigate whether more flexible LLMs as relatively unbiased language learners \cite{wilcox2023using} are appropriate tools to study how languages evolve. While their internal mechanisms are fundamentally different from humans, they still are the first close flexible comparators of human language users that can be used as tools to answer cognitive and typological investigations \cite{warstadt2022artificial, van-dijk-etal-2023-large}. Given that languages are shaped by the biases and pressures of individual language learners, which are different for LLMs (e.g., fewer memory constraints), we are interested in finding similarities and differences between humans and LLMs on specific language evolution-oriented tasks. 

Our work largely follows the experimental design by \citet{Kirby2015CompressionStructure} in which Bayesian learners and humans learn an artificial language to communicate in a referential game. They find that linguistic structure arises from a trade-off between pressures for compressibility and expressivity. Our work extends their work by using LLMs as objects of investigation. Specifically, we investigate how artificial languages evolve when two LLMs communicate in a referential game and what the effects of generational transmission on these languages are. We compare properties of these languages to those that are found in experiments involving humans. Results show that 1) LLMs can learn artificial languages and use them to communicate successfully, 2) the languages exhibit higher degrees of structure after multiple communication rounds, 3) LLMs generalise in more systematic ways when the evolved language is more structured, and 4) languages adapt, although not necessarily in a human-like way, and become easier to learn by the LLMs as a result of generational transmission.

\section{Background \& Related work}
\subsection{The evolution of structure}
Learning novel signal-meaning mappings, and the emergence of rules that can combine these signals into structured languages have been abundantly investigated in the field of language evolution using human experiments \cite{kirby2008cumulative, galantucci2005experimental, scottphillips2009signalling, verhoef2012origins, raviv2019compositional,raviv2019larger, kouwenhoven2022need} and computational simulations \cite{deBoer2006computer, steels2012grounded, lazaridou2020emergent, kouwenhoven-etal-2024-curious}. These typically follow a setup where success depends on cooperation between two or more participants/agents in a Lewis game. Here, players are prevented from communicating using conventional communicative means and instead must establish novel communication systems through repeated cooperation. Outcomes often show that players, human or machine, quickly establish novel signal-meaning mappings that enable them to communicate successfully. However, recent computational simulations using reinforcement learning agents often develop communicative systems different from those of humans \cite{galke2022emergent}\footnote[1]{But see \citet{lian2023communication, lian2024NellCom-X, zhang-etal-2024-endowing} for recent work showing that the need to be understood (i.e. communicative success), noise, context sensitivity, and incremental sentence processing help induce human-like patterns of dependency length minimisation in reinforcement learning agents.} unless specific key pressures are introduced to recover initially absent human patterns \cite{galke2024emergentcommunicationlearningpressures}. 

It has been suggested that seemingly arbitrary aspects of linguistic structure may result from general learning and processing biases deriving from the structure of thought processes, perceptuo-motor factors, cognitive limitations, and pragmatics \cite{Christiansen2008LanguageBrain}. A well-investigated cause for this phenomenon is the process of cumulative cultural evolution \cite{boyd1996culture, tomasello1999cultural}, which is typically investigated using iterated learning experiments \cite{kirby2008cumulative}. Here, information (e.g., a language) is repeatedly passed down from one generation to the next, where the information is modified and improved upon within each generation. The influential work from \citet{kirby2008cumulative, Kirby2015CompressionStructure} shows that when human individuals learn an artificial language that was previously learned by another individual, languages become easier to learn and display a higher degree of structure. Crucially, these results are mostly attributed to the fact that the language repeatedly goes through a learning bottleneck, in which individual cognitive biases such as memory constraints slowly shape the language. Iterated learning has been used to show that structure emerges in various setups with, for example, continuous signals \cite{verhoef2012origins} or continuous meaning spaces \cite{carr2017cultural}, and it is argued that it may have led to the statistical Zipfian structure of language \cite{Arnon2024CulturalLanguage}. Yet, \citet{raviv2019compositional} showed that structure can also emerge \textit{without} generational transmission. In this case, a pressure for compressibility originating from communication with multiple interaction partners and expanding meaning spaces causes languages to become compositional. This effect is even more prominent if the number of interaction partners is larger \cite{raviv2019larger}. The current work is inspired by the traditional methods described before and extends them with our current most sophisticated models of natural language. 

\begin{figure*}[ht]
    \centering
    \includegraphics[width=\linewidth]{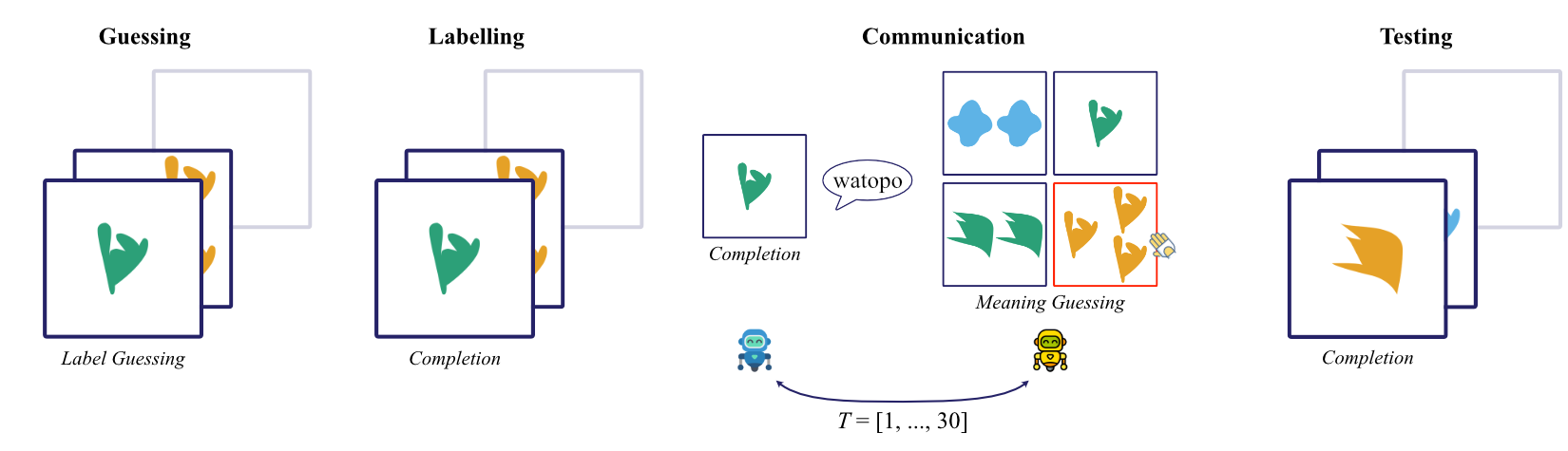}
    \caption{A graphical representation of the experimental blocks. The agents first go through a guessing block before labelling each of the 15 training stimuli in the labelling block. The communication block is done for 4 rounds each consisting of 30 tasks $T$, where the agents alternate speaker-listener roles to be a speaker and listener for each stimulus once. Finally, the agents label 27 (15 original and 12 novel) stimuli in the testing block.}
    \label{fig:overview}
\end{figure*}

\subsection{LLMs as models of language}
LLMs are sophisticated models of natural languages and growing evidence shows their ability to exhibit `average' human behaviours. It is, for example, suggested that LLMs can model human moral judgements \cite{Dillion2023CanParticipants} and transmission chain experiments revealed human-like content biases in GPT-3.5 \cite{Acerbi2023LargeExperiments}. 
% But contrarily, human-like response biases are not found in survey questions across all LLMs \cite{tjuatja2024responsebiases}. 
When LLMs are extended with records of experiences, \citet{park2023generative} showed that groups of generative agents display believable human-like individual and emergent social behaviours when they interact over extended periods. It is even suggested that human-LLM interactions in everyday life can potentially mediate human cultures through their influence on cultural evolutionary processes of variation, transmission and selection \cite{brinkmann2023machine, Yiu2024immitationinnovation}. 

While previous work has investigated human-like behaviour at inference time, findings from cognitive science can also be used to improve model performance. Iterated learning can, for example, be incorporated into the training regime to extrapolate desirable behaviours. \citet{zheng2024iterated} have likewise shown that representations are easier to learn when vision-language contrastive learning is reframed as the Lewis signaling game between a vision agent and a language agent, ultimately improving compositional reasoning in vision-language models. However, this does not guarantee model improvements, \citet{shumailov2024ai} have for example shown that LLMs, autoencoders and Gaussian mixture models drift when trained repeatedly on AI-generated data. In these cases, crucially, the generated content is slowly optimised to be understandable for models, \textit{not} for humans, resulting in what they call model collapse. The authors therefore argue that genuine human interactions with systems will be increasingly important to prevent model collapse. While drift is often seen as an unwanted effect of unsupervised training, from a language evolution viewpoint this is not surprising, since languages adapt to how they are learned and used. As such, it has been suggested that languages should adapt to become more natural for humans \textit{and} machines \cite{kouwenhoven2022emerging} and that findings from cognitive science can prevent modal collapse \cite{smith2024ai} or inform modelling choices \cite{galke2024emergentcommunicationlearningpressures}. Here, we view LLMs from this evolutionary perspective.

Although biases inherent to a language model's (pre-)training objectives (i.e. the cloze task and instruction tuning) and memory constraints are very different from those in humans, recent work has shown that GPT-2 models struggle to learn languages that contain unnatural word orders, lack hierarchical structure, or lack information locality \cite{kallini-etal-2024-mission}. This suggests that, even though the language processing mechanisms in transformers are non-humanlike, LLMs share some preference for structured languages similar to humans. Moreover, in an artificial language learning experiment similar to the work presented here, \citet{Galke2023WhatMakes} showed that compositional structure is advantageous for GPT-3 when learning an artificial language and that a higher degree of compositional structure also resulted in human-like generalisation for new unseen items. Our work is different in that \citeauthor{Galke2023WhatMakes} tested the ability of GPT-3 to learn languages that evolved during a \textit{human} experiment \citep{raviv2019larger, raviv2021easytolearn}, thus being optimised for human learners. We instead wish to investigate what kinds of languages evolve when they are optimised for \textit{LLMs}.  

\begin{prompt*}[htbp]
\small
\begin{lstlisting}[mathescape=true,escapeinside={*@}{@*}]
             {'shape':3,'colour':'blue','amount':1,'word':'ninikonu'}
             {'shape':1,'colour':'green','amount':3,'word':'hanosa'}*@\vspace{.4em}@*
                 $\smash\vdots$
             {'shape':2,'colour':'orange','amount':2,'word':'sanu'}
             {'shape':1,'colour':'green','amount':3,'word':'[COMPLETE]
\end{lstlisting}
\caption{A vocabulary snippet used in a completion prompt. Full prompts are visible in Appendix \ref{sec:prompts}}\label{fig:cmpl-prompt}
\end{prompt*}

\section{Methodology}
Our methodology is inspired by \citet{Kirby2015CompressionStructure} and \cite{raviv2021easytolearn}. The complete simulation set-up consists of four blocks: guessing, labelling, communication and testing (§\ref{sec:blocks} \& Figure \ref{fig:overview}\footnote{This is for illustration purposes only, we stress that our simulations are entirely run in the textual modality only to avoid the additional challenge of extracting relevant visual features and mapping these to artificial languages.}). The agents perform the guessing, labelling, and testing block separately; communication is interactive. The communication block is a classic referential game in which two agents communicate to discriminate a target stimulus from four distractor stimuli. They do so in four rounds, each consisting of 30 interactions $T$, alternating speaker-listener roles between interactions. In a single interaction round, the speaker observes a target stimulus and utters a signal that describes the current stimulus. Using this utterance, the listener must discriminate the correct target. Cooperation is successful when the listener's guess is the target stimulus.

\subsection{Stimuli and initial languages}
\label{stimuli-languages}
The meaning space consists of stimuli with three attributes. They have one of three shapes, one of three colours, and can appear in groups of one, two, or three shapes, creating 27 distinct stimuli. Initial signals for these stimuli were generated before each experiment according to the method used by \citet{kirby2008cumulative}. The signals are concatenations of 2, 3, or 4 randomly picked consonant-vowel (CV) syllables resulting in artificial non-existing signals (e.g., watopo, nafa, nomomeme). The CV syllables consist of one of eight consonants g, h, k, l, m, n, p, w and one of five vowels a, e, i, o, u. Out of 27 stimuli, only 15 stimuli are used during the guessing, labelling, and communication blocks. All 27 stimuli are used in the testing block. The training stimuli are selected randomly before each simulation, but we ensure that each attribute value is represented equally often across this set.

\subsection{Simulation blocks}
\label{sec:blocks}
Each simulation consists of four blocks. In the first block, we assess whether agents can guess the right signal when presented with a stimulus. Second, in the labelling block, an agent repeatedly produces a signal for each stimulus given the initial training vocabulary. The signals generated in this block are taken as the learned vocabulary for that agent. In the third block, the agents communicate as described before, taking turns as speaker and listener until all rounds are completed and each stimulus appeared twice per round (i.e., both agents produced a signal for each stimulus and made a guess for each stimulus). In this block, the interaction between the agents slowly alters each agent's individual vocabulary much like done by \cite{deboer2000self-oganization, steels2012grounded} by updating the current stimulus to be associated with the produced signal. After the communication block, the testing block tasks the agents to generate signals for the entire meaning space of 27 stimuli using the training vocabulary that was optimised in the labelling and communication block. Hence, they must generalise their strategies to unseen samples.

\subsection{LLMs as agents}
The LLMs in our experiment were instruction-tuned instantiations of Llama 3 70B \cite{dubey2024llama3herdmodels} with greedy sampling\footnote{Although we only report results on one model type, initial explorations with GPT-3.5 and Llama 2 7B showed similar behaviours to LLama 3 70B}. While human participants typically learn signal-meaning mappings through a learning block, we use LLMs' in-context learning \cite{brown2020few} ability to teach them the languages. Specifically, we prepend our prompts with the items to be learned in a structured JSON-like format (Prompt \ref{fig:cmpl-prompt}). Given the observed behavioural similarities between humans and LLMs \cite{Galke2023WhatMakes}, we assume that a vocabulary of signal-meaning mappings in the context of a prompt provides enough (distributional) information for a LLM to learn an appropriate mapping between the attributes of the stimuli and signal syllables. Although the prompt structure `invites' the LLM to infer a signal from the stimulus attributes, we are agnostic about how exactly and what kind of mapping the LLM deduces, but we are interested in the resulting behaviours. 

Throughout a simulation, agents essentially perform one of two tasks: generation or guessing. The labelling block and speaking in the communication block involve generating signals. The guessing block and discrimination in the communication block involve guessing. The prompts for these tasks are extensions of those used by \citet{Galke2023WhatMakes}, with slight adaptations to enable LLMs to discriminate between stimuli. Given that LLMs show a primacy and recency bias \cite{liu-etal-2024-lost}, the vocabulary is shuffled before each task such that ordering effects are minimal. System instructions depend on the task performed but are largely similar and chosen to be maximally close to instructions given to humans in experimental settings. 

\textbf{Generating signals.} For signal generation in the labelling block, we use prompt completion (Prompt \ref{fig:full-cmpl-prompt-labelling}). During labelling, the agents see the \textit{entire} training set and generate a signal for each stimulus, effectively amounting to a look-up task since the stimulus is present in the prompt. On the other hand, the vocabulary presented to agents during communication and testing does \textit{not} include the current stimulus, thus requiring the agents to extract an appropriate mapping and generalise to new stimuli (Prompt \ref{fig:full-discr-prompt-speaker}). A human-like solution would be to map stimulus attributes (i.e. shape, colour, and amount) to syllables representing these attributes and create compositions that describe the stimulus. During communication, we add a \texttt{communicativeSuccess} attribute which is set to $1$ if the previous interaction for this stimulus was successful and zero otherwise. Adding this attribute functions as a memory between interactions and provides a pressure for expressivity. It is hypothesised that the latter plays an important role in human language evolution since it prevents languages from becoming degenerate \cite{smith2013linguistic}. Importantly, during testing, the vocabulary presented to the agents always includes the train set (without the current stimulus), and items from the test set are never present.

\textbf{Guessing signals or meanings.} For guessing and discrimination during communication, the agents need to respond with a choice corresponding to the speaker's signal. Unfortunately, LLMs are inconsistent and unreliable in answering multiple choice questions \cite{khatun2024study}. In our initial exploration, this indeed proved to be unusable. Instead, for each distractor (signal or meaning), we run the prompt prefilled with that distractor through the model and select the distractor with the highest probability (Prompt \ref{fig:full-discr-prompt-listener}). Again, the agents observe the training vocabulary \textit{with} the current stimulus in the guessing block. In the communication block, agents observe the training vocabulary \textit{without} the current stimulus.

\subsection{Metrics}
We are firstly interested in investigating whether two agents settle on a language that enables them to communicate, measured by the percentage of successful interactions (\textit{PercCom}) in a round. We use multiple metrics to measure structure in messages. The most common metric is topographic similarity \cite[\textit{TopSim},][]{Henry2006Understanding}. Similar to \citet{kirby2008cumulative}, we report Z-scores of the Mantel test \cite{mantel1967detection} between signal similarities (normalised Levenshteins distance) and semantic similarities (the number of equal attributes between two meanings). A communication system with a high \textit{TopSim} uses similar signals for similar meanings. We compute the \textit{Ngram} diversity \cite{meister-etal-2023-locally}, being the average fraction of unique vs. total \textit{Ngrams} for $N \in \{1, 2, 3, 4, 5\}$ in all produced signals. Low \textit{Ngram} diversity across all signals implies the agents re-use parts of signals in different signals, hinting at compositional signals when it happens in combination with increased \textit{TopSim}.
We assess the degree of signal systematicity between the signals produced for unseen stimuli in the test block and the previous stimuli in the communication block using the generalisation score \cite[\textit{GenScore},][]{raviv2021easytolearn}. Here, we first compute the pairwise semantic difference between each stimulus in the train and test scenes, followed by the pairwise normalised edit distance between the signals produced for these scenes. We then take the Pearson correlation between these differences across all stimuli. Intuitively, this measures whether similar scenes across both sets are similarly labelled, thereby suggesting generalisation. 

\begin{figure*}[ht]
    \centering
    \begin{subfigure}[t]{1\columnwidth}
        \centering
        \includegraphics[width=0.95\columnwidth]{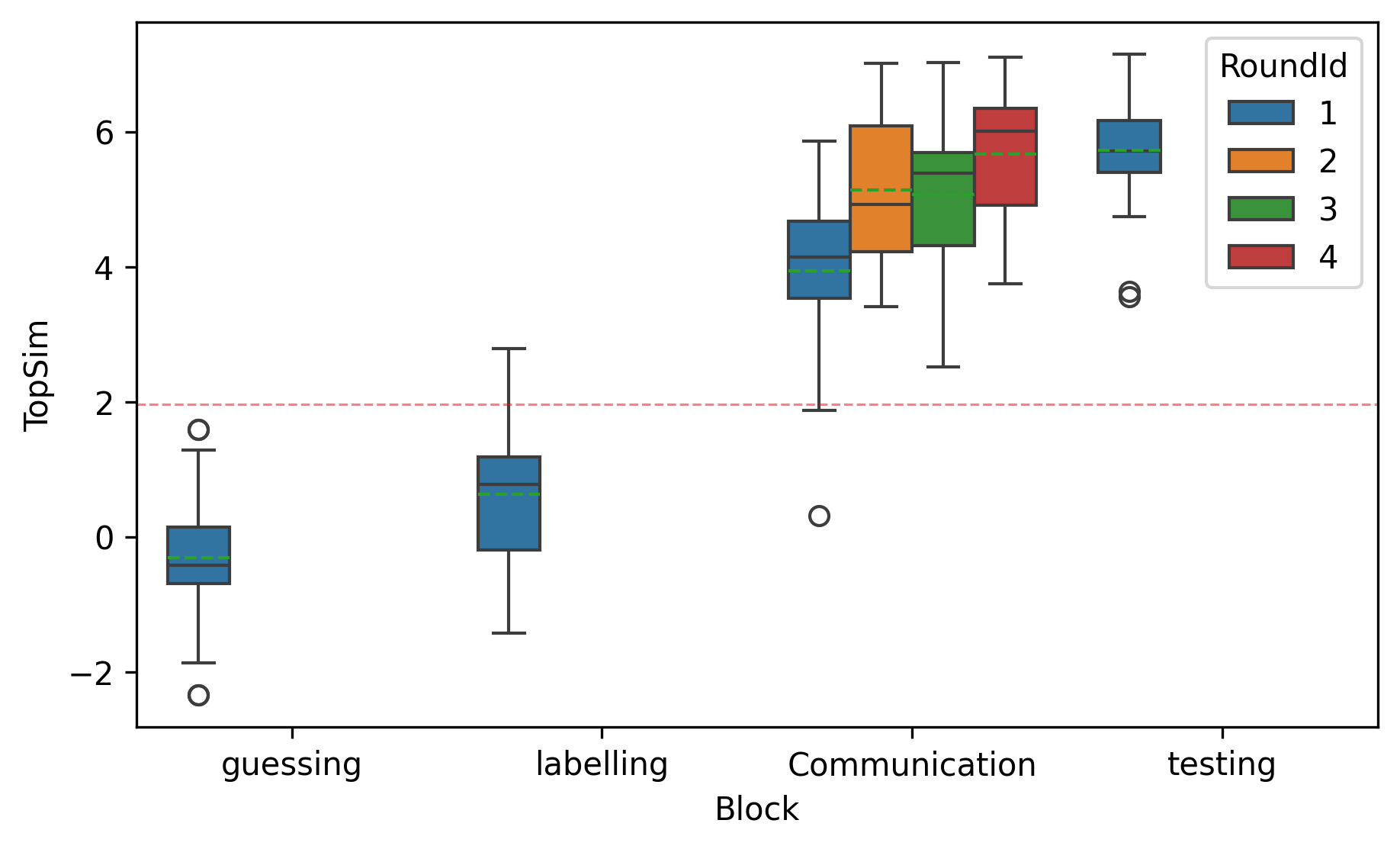}
        \caption{TopSim scores over the agent's vocabulary in each block and round. The dashed red line indicates the $p<.05$ level.}
        \label{fig:combined-TopSim}
    \end{subfigure} 
    \hfill
    \begin{subfigure}[t]{1\columnwidth}
        \centering
        \includegraphics[width=0.95\columnwidth]{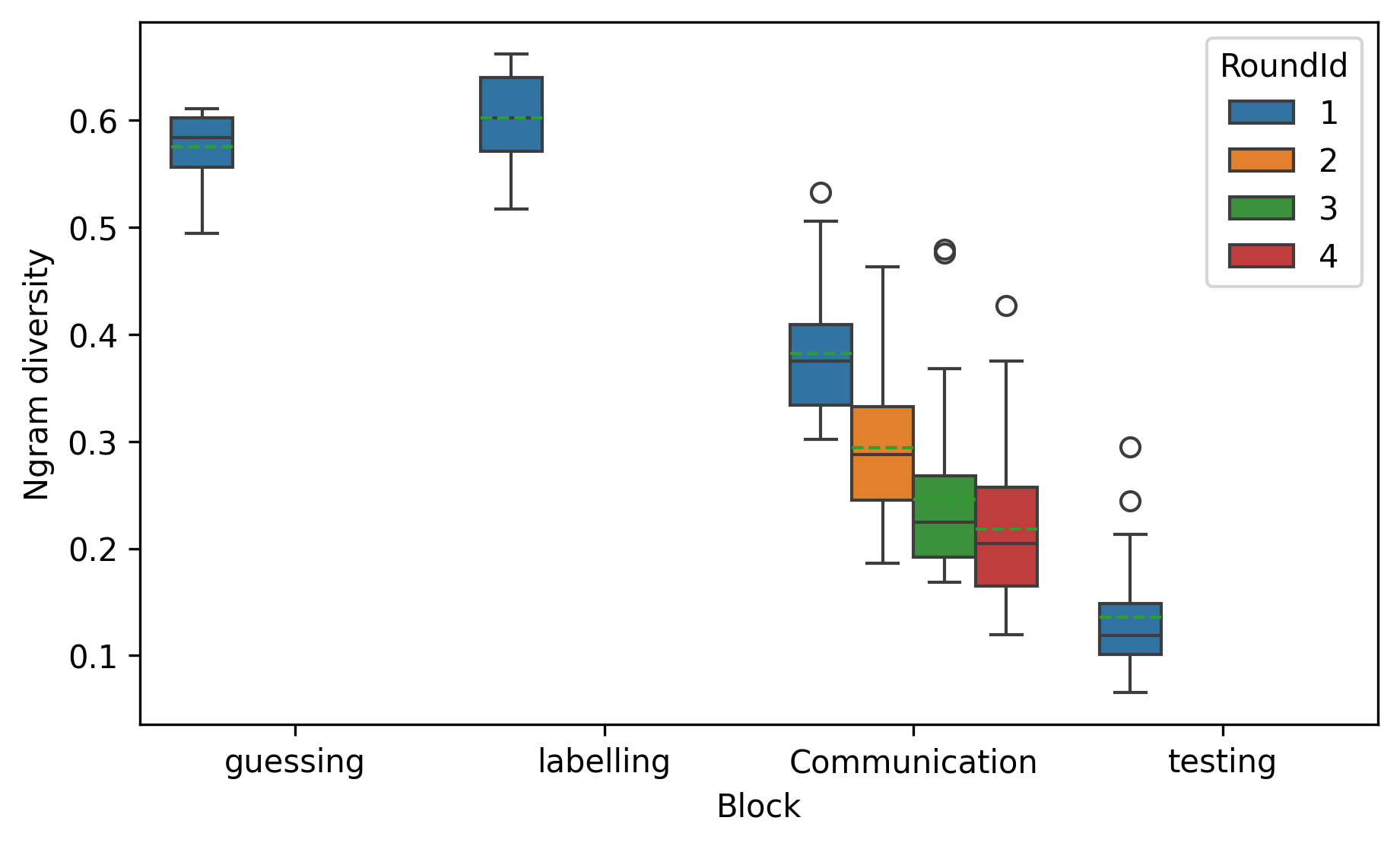}
        \caption{Ngram diversity scores over the agent's vocabulary in each block and round.}
        \label{fig:combined-Ngram}
    \end{subfigure}
    \caption{Communication clearly increases the structure of the vocabularies, as seen by the increasing \textit{TopSim} scores and decreasing \textit{Ngram} diversity.}
    \label{fig:overall-results}
\end{figure*} 

\section{Evaluation}
We ran 15 simulations, each initialised with a random seed and unique artificial unstructured language. Metrics were computed for each block, except for the generalisation score, which is only computed for the testing block. A human-like result would show increasingly successful interactions and increasing \textit{TopSim} scores, while \textit{Ngram} diversity should go down. If this is the case, we expect to observe higher generalisation scores since agents can compose new signals according to a learned structured strategy. We use linear mixed effects models to analyse the results of the communication block to take the random effects of each simulation's vocabulary into account. The slope ($\hat{\beta}$) determines the direction of the effect and the rate of change. Additionally, we use conditional $R^2$ \cite{nakagawa2013r2}, denoted by $R^2_c$, which considers fixed and random effects, to show how much variance can be explained by the model. Higher values of $R^2_c$ indicate that the model captures more variance and that correlations are stronger. Finally, we report the marginal $R^2_{m}$, which is the variance explained by the fixed effects.

\section{Results}
\subsection{Learning the artificial languages}
We first assess whether LLMs were able to learn the initially unstructured languages. Given the nature of the guessing task, which is essentially a lookup task, unsurprisingly, LLMs were able to guess the correct signals for the stimuli almost perfectly ($M=.973, SD=.031$). However, labelling the same stimuli via completion proved much more difficult ($M=.453, SD=.152$) despite the presence of the correct signal in the prompt. This contrast is in line with work showing that LLM predictions are sensitive to task instructions and how predictions are extracted \cite{weber-etal-2023-mind, hu-levy-2023-prompting, hu2024auxiliary}. Additionally, it corroborates using prefilled options in our guessing prompts during communication. Nevertheless, this performance is still better than that of humans\footnote{Preliminary analyses of an ongoing experiment involving humans show that the guessing block is much easier than the labelling block.} and is not unimpressive given the vast number of possible signals that can be produced. Finally, the expected struggle to correctly reproduce (i.e., learn) unstructured signals introduces some welcome variation to the agents' vocabulary which is used at the start of the communication block.

\subsection{Agents communicate successfully}
Once the agents have individually learned the vocabulary, they start communicating. Despite initially starting with different languages, approximately $70\%$ of the interactions in the first round are successful (chance performance would amount to $25\%$). This increases somewhat in the following rounds to $\approx75\%$, but not significantly (Appendix \ref{sec:percCom}, Figure \ref{fig:percCom}). Interestingly, communicative success is not guaranteed, it fluctuates between rounds and in some simulations it even decreases drastically.

\subsection{Communication results in structure}
Although the initial languages are unstructured, some form of structure emerges due to repeated learning and use (Figure \ref{fig:overall-results}). This mostly happens during the communication block where \textit{TopSim} increases significantly across rounds ($\hat{\beta}=.508\pm.073, R^2_{c}=.579, R^2_{m}=.355, p<.001$) and \textit{Ngram} decreases across rounds ($\hat{\beta}=-.054\pm.004, R^2_{c}=.812, R^2_{m}=.558, p<.001$). This increase in structure benefits communicative success positively ($\hat{\beta}=.035\pm.007, R^2_{c}=.769, R^2_{m}=.427, p<.001$). However, we also observe behaviour that is not human-like; the signals used to communicate become longer over the rounds ($\hat{\beta}=.557\pm.044, R^2_{c}=.919, R^2_{m}=.505, p<.001$). This contradicts what is observed in human experiments, where we typically observe that messages become shorter and lie close to a theoretical frontier balancing expressivity and simplicity \cite{piantadosi2011wordlengths, Kirby2015CompressionStructure}. 

These results extend the findings of \citet{Galke2023WhatMakes}; LLMs not only learn structured vocabularies better but also naturally shape languages to have some form of structure when they are optimised for their inherent biases. As LLMs struggle to learn impossible languages \cite{kallini-etal-2024-mission}, reframing prompt instructions into a structured list improves the model response \cite{mishra-etal-2022-reframing}, and given that we do not impose pressure to induce structure, the surprising outcome of our experiments may be the result of an apparent ``structure bias'' in LLMs.

\begin{figure}[!t]
    \centering
    \includegraphics[width=1\columnwidth]{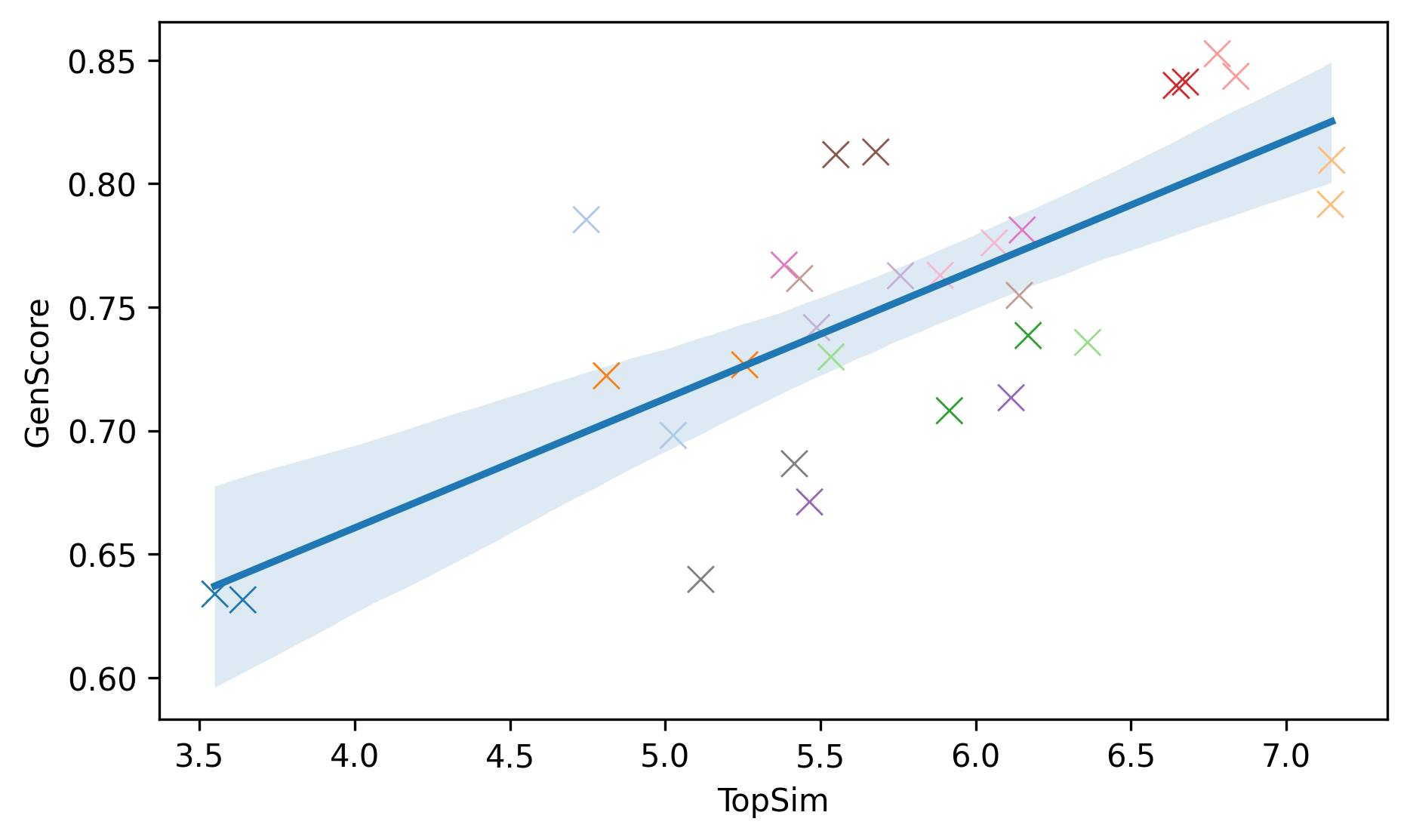}
    \caption{Languages that have evolved to be more structured allow for better generalisation to unseen test stimuli. Colours refer to individual simulations.}
    \label{fig:topsim-genscore}
\end{figure}

\subsection{Structure enables better generalisation}
After the communication block, the agents engage in the final simulation block. Here they generate signals for all 27 stimuli using the vocabulary that has evolved after learning and communication. We find that high \textit{TopSim} languages allow for better generalisation ($r=0.735, p<.001$, Figure \ref{fig:topsim-genscore}). 

A qualitative inspection of the signals generated in the testing block of the simulation which resulted in the highest \textit{TopSim} after communication reveals that this agent repeatedly re-uses parts of signals in different compositions (Table \ref{tab:example_generalisation}). For example: ``su'' refers to the amount one, ``pepi'' to two, ``petite'' to three. For shape 1, the signals ``sunu'' and ``sutu'' are used, ``ginu'' for shape 2, and shape 3 is referred to with ``wipi'' or ``wipu''. However, colours are less clearly demarcated by unique signal parts. This is also reflected in the ratio of unique signals produced during the test block ($M=62.1\%, SD=19.8\%$), showing that some simulations sometimes result in repetitive use of the same signals for different meanings, resulting in a somewhat degenerate vocabulary. Nevertheless, it is clear that unseen stimuli are often labelled similarly to previously seen stimuli. 

\begin{table}[!ht]
    \centering
    \begin{tabular}{c|c|l|c|l}
        \textbf{} & \textbf{Shape}   & \textbf{Colour} & \textbf{Amount} & \textbf{Word}\\
        \hline \hline
        \parbox{1em}{\multirow{15}{*}{\rotatebox{90}{train set}}}   & 3 & orange    & 1 & wipisu\\
                                                                    & 1 & green     & 2 & sutupepi\\
                                                                    & 2 & green     & 1 & ginisu\\
                                                                    & 3 & green     & 1 & wipisu\\
                                                                    & 1 & blue      & 2 & sunupepi\\
                                                                    & 1 & green     & 3 & sutupitite\\
                                                                    & 2 & orange    & 1 & ginusu\\
                                                                    & 3 & blue      & 3 & wipipitite\\
                                                                    & 3 & green     & 3 & wipupitite\\
                                                                    & 3 & blue      & 1 & wipisu\\
                                                                    & 1 & blue      & 3 & sunupitite\\
                                                                    & 2 & orange    & 3 & ginupitite\\
                                                                    & 2 & blue      & 2 & ginupepi\\
                                                                    & 1 & orange    & 2 & sunupepi\\
                                                                    & 2 & orange    & 2 & ginupepi\\
        \hline \hline
        \parbox{1em}{\multirow{12}{*}{\rotatebox{90}{test set}}}    & 1 & orange    & 1 & sutisu\\
                                                                    & 1 & orange    & 3 & sutupitite\\
                                                                    & 1 & green     & 1 & sutusu\\
                                                                    & 1 & blue      & 1 & sunusi\\
                                                                    & 2 & green     & 2 & ginupepi\\
                                                                    & 2 & green     & 3 & ginupitite\\
                                                                    & 2 & blue      & 1 & ginisu\\
                                                                    & 2 & blue      & 3 & ginupitite\\
                                                                    & 3 & orange    & 2 & wipupepi\\
                                                                    & 3 & orange    & 3 & wipipitite\\
                                                                    & 3 & green     & 2 & wipupepi\\
                                                                    & 3 & blue      & 2 & wipupepi\\
    \end{tabular}
    \caption{The signals produced in the testing phase of the simulation that resulted in the highest \textit{TopSim} score (7.13) after communication. The signals for the test stimuli share parts of signals and are composed similarly to train stimuli ($GenScore=.792$)}
    \label{tab:example_generalisation}
\end{table}

\section{Iterated learning}
The previous results showed that two LLMs can successfully communicate and slowly shape the language to become more structured. Provided that cumulative cultural evolution can extrapolate weak biases to have strong effects in socially learned systems like language \cite{smith2011learning}, we extend our simulations by adding generations of learners. The first generation is initialised with a random unstructured language described in Section \ref{stimuli-languages}, but in following generations, agents learn a portion of the signal-meaning mappings produced in the testing block by the agents of the previous generation. Only the vocabulary of the agent with the highest \textit{TopSim} is transmitted to the next generation. We ran six transmission chains of 8 generations each. The seed generations for each chain were selected randomly from our initial 15 simulations. 

\subsection{Learnability increases}
\begin{figure}[t]
    \centering
    \includegraphics[width=1\linewidth]{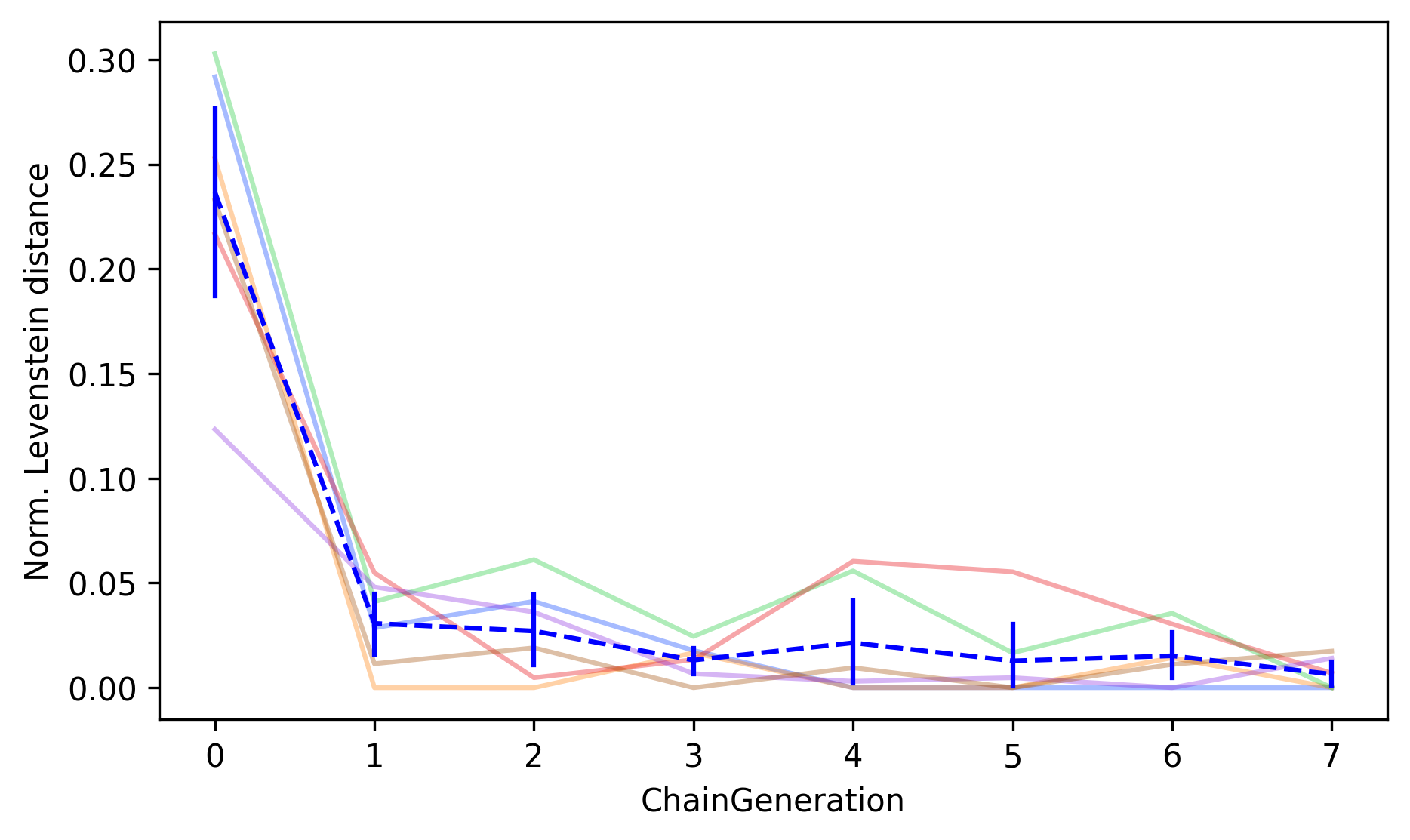}
    \caption{The normalised Levenshtein distance between the ground truth and the produced signal in the learning block. Solid lines indicate chains, dashed blue line indicates the average Levenshtein distance across chains.}
    \label{fig:labelling-chain}
\end{figure}

Figure \ref{fig:labelling-chain} clearly reveals that the learnability increases. While LLMs in the first generation struggle to look up signals and reproduce them, a single generation learning and using a language tremendously decreases the edit distance between ground truth signals and the produced signals. These results are remarkably similar to findings with human participants \cite{Kirby2015CompressionStructure}, and show that the languages are optimised for LLMs' preferences. 

\subsection{Communicative success and non-humanlike structures}
Despite the increase in learnability, we do not observe an increase in communicative success due to iterated learning (Appendix \ref{sec:iterated-learning} Figure \ref{fig:chain-PercCom}). Possibly, this is due to the already high scores of the first generation. Despite their increased learnability, the signals become significantly longer and more ambiguous. We take this non-humanlike solution to be an artefact of an absence of pressures for memorisation in LLMs'. While human language is optimised to be compressible and expressive \cite{fedzechkina2012langauge, tamariz2015culture, Kirby2015CompressionStructure}, context windows of LLMs, are considerably larger. In our case, Llama 3 70B has a context window of 8.2K tokens, which we do not exceed. 

Finally, the metrics to measure structure display a mixed picture. \textit{TopSim}, does increase but not significantly across generations (Figure \ref{fig:Structure-Chain} and Table \ref{tab:ttests} in appendix \ref{sec:iterated-learning}) but \textit{Ngram} diversity decreases significantly across generations. Qualitative inspections of several vocabularies show that some languages evolve into degenerate languages with repeating signals for different stimuli (i.e., underspecification). Corroborated by a significantly lower number of uniquely produced signals in the last generation compared to the first simulation ($t(5)=2.64, p=.046, M_{gen0}=.707, SD_{gen0}=.142, M_{gen7}=.519, SD_{gen7}=.119$). Together this causes the \textit{Ngram} diversity to be lower while clearly hurting communicative expressiveness. Even though degenerate languages are not uncommon in iterated learning experiments with humans \cite[e.g., experiment 1 in][]{kirby2008cumulative}, an additional pressure for expressivity typically prevents languages from becoming underspecified. Given the expressivity pressure that we expect to result from the communication block, we expected to see less underspecification. Iterated learning therefore results in vocabularies optimised for LLM agents but do so in a non-humanlike way. 

\section{Discussion}
Our findings show a mixed picture, agents comprised of LLMs can learn and use artificial languages in a referential game. They do so by optimising the initially holistic vocabulary to fit better with the preferences of their language model, resulting in increased regularity and structure (Table \ref{tab:example_generalisation}). These human-like results are much in line with previous findings showing that structured languages can emerge from repeated interactions between interlocutors \cite[i.a.][]{Selten2007emergence,verhoef2016cognitive,nolle2018emergence,raviv2019compositional}. Yet, we also observe some degeneracy, i.e. many-to-one mappings of signals and attributes, and non-humanlike behaviours such as a tendency to produce long signals. Iterated learning further increases the learnability of the vocabulary but also extrapolates these non-humanlike behaviours further. Despite not being able to \textit{directly} compare our results to human data, these findings are loosely comparable to earlier work involving human participants \cite{Kirby2015CompressionStructure,raviv2019larger} in which languages with similar properties emerge. 
% This also raises questions regarding the commonly used \textit{TopSim} metric, which numerically increases but clearly does not take degeneracy into account. 

Table \ref{tab:example_generalisation} suggests that certain attributes, such as the colour attribute, in the inputs may be ignored, possibly due to the primacy and recency bias in LLMs \cite{liu-etal-2024-lost}. Optimising the instructive sentences by choosing sentences that maximise the fraction of valid model answers for each task, as suggested by \citet{aher2023using}, may alleviate these ignorances and increase focus on relevant attributes. It is also possible that the LLMs do not `experience' enough pressure to be understood by other agents, i.e., the \textit{communicativeSuccess} attribute is not able to force a need to be expressive, which is deemed an essential pressure in computational simulations for human-like structures \cite{galke2024emergentcommunicationlearningpressures}. Despite these discrepancies, it is nevertheless interesting that some form of structure emerges. 

Our results further show variability between generations of learners. This is not uncommon in human experiments where processes of interaction and transmission sometimes generate fully systematic, compositional languages, but can also result in systems that lack structure entirely \cite{verhoef2022interaction}. Differences in personal biases may be a contributing factor to these differences \cite{kouwenhoven2022need}. Since we do not initialise agents with different biases, these variations, originating in distributional information of the prepended vocabularies, are a natural human-like outcome of repeated exposure to and use of the language.

The evolution of degenerate vocabularies could be explained by the use of greedy decoding during signal generation, which does not necessarily produce the most human-like text \cite{holtzman2020thecurious, meister-etal-2022-high, meister-etal-2023-locally} and may therefore also result in non-humanlike composition. Moreover, once an agent, perhaps mistakingly, duplicates a signal, its raw probabilities are increased when producing the next utterance, possibly resulting in a feedback loop that collapses onto a degenerate vocabulary. This effect may be further increased due to LLMs' inability to innovate \cite{bender2021stochastic, Yiu2024immitationinnovation} and the choice of structured prompts that do not explicitly ask for innovation. Future work could attempt to increase the composition of novel signals by increasing the temperature parameter. Perhaps resulting in slightly more novel outputs as this forces exploration of the vocabulary embedding space \cite{peeperkorn2024temperature}, possibly alleviating the evolution of degenerate vocabularies and shifting the optimisation of the language to different solutions. 

The rapid increase in learnability resulting from iterated learning proves that weak learning biases in language models, such as, for example, an observed simplicity bias \cite{chen2024sudden}, can be amplified by the process of generational transmission. Additional simulations with increased communicative difficulty, e.g., by increasing the number of distractors or the number of interaction partners, could reveal whether and how some form of memory constraint affects the learnability of the languages. Doing so additionally captures the diversity and dynamic nature of language in the real world. In general, systematic manipulations across model features (e.g., size, training data, or decoding strategies) may expose why we observe tendencies such as producing longer signals. Similar to what was proposed by \citet{galke2024emergentcommunicationlearningpressures}, we argue that careful manipulation of our setup can help reveal underlying mechanistic biases of language models and inform modelling choices when simulating language acquisition in LLMs. Taking into account the important role communication plays in shaping human language, LLM performance drastically increased when it was optimised for successful communication through reinforcement learning from human feedback (RLHF).

Finally, we acknowledge that our results depend on many methodological considerations, such as the prompt format, task instructions, and the tokenisation process. However, our primary goal was to investigate whether LLMs can be used in simulations of artificial language emergence. We aimed to stay maximally close to well-known experimental methods in the field of language emergence and did not optimise for performance, human-like results, or compositional vocabularies. Instead, our goal was to reveal LLMs' natural behaviours resulting from learning and using artificial languages. Future work could extend our findings by performing experiments in which humans collaborate with LLMs to investigate whether languages can evolve that are optimised for human \textit{and} LLM preferences. 

\section{Conclusion}
Given the remarkable linguistic abilities of recent LLMs, we show how they behave in a classical referential game in which artificial languages, typically used in the field of language evolution, are learned and used. Primarily, our results suggest that LLMs can be used as artificial language learners to investigate the evolution of language. We show that initially unstructured languages are optimised for improved learnability and allow for successful communication. While we found some evidence of human-like compositional structures that enhance generalization abilities, we also identified notable differences in behavioural characteristics of LLMs in comparison to humans. Notably, iterated learning processes increased vocabulary learnability but also amplified such different characteristics further. As such, we extend existing research by revealing that structured languages are not merely easier for LLMs to learn. Critically, the inherent biases of LLMs also shape unstructured languages towards increased regularity. These findings contribute to a deeper understanding of how LLMs process and evolve language, potentially bridging the gap between computational models and natural language evolution. Finally, we hope to have shown that our setup is useful in exposing the underlying mechanistic biases of LLMs and demystifying their uninterpretable nature.

\section*{Acknowledgments}
We wish to thank Bram van Dijk for his comments on an early draft of this paper and the helpful discussions.  

\bibliography{anthology,custom}

\newpage
\appendix

\begin{table*}[ht]
\centering
\renewcommand{\arraystretch}{1.2}
\begin{tabular}{lcccccc}
\hline
& $t(5)$ & $p$ & $M_{gen0}$ & $SD_{gen0}$ & $M_{gen7}$ & $SD_{gen7}$ \\
\hline
\textit{TopSim} & -1.42 & .215 &  9.62 & 1.21 & 10.5 & 1.77\\
\textit{Ngram}  & 2.83 & .037 & .158 & .074 & .071 & .025\\
\hline
\end{tabular}
\caption{Paired t-tests shows that \textit{Ngram} does significantly change resulting from generational transmission, while \textit{TopSim} does not.}
\label{tab:ttests}
\end{table*}

\section{Communication per round}
\label{sec:percCom}

\begin{figure}[ht]
    \centering
    \includegraphics[width=1\linewidth]{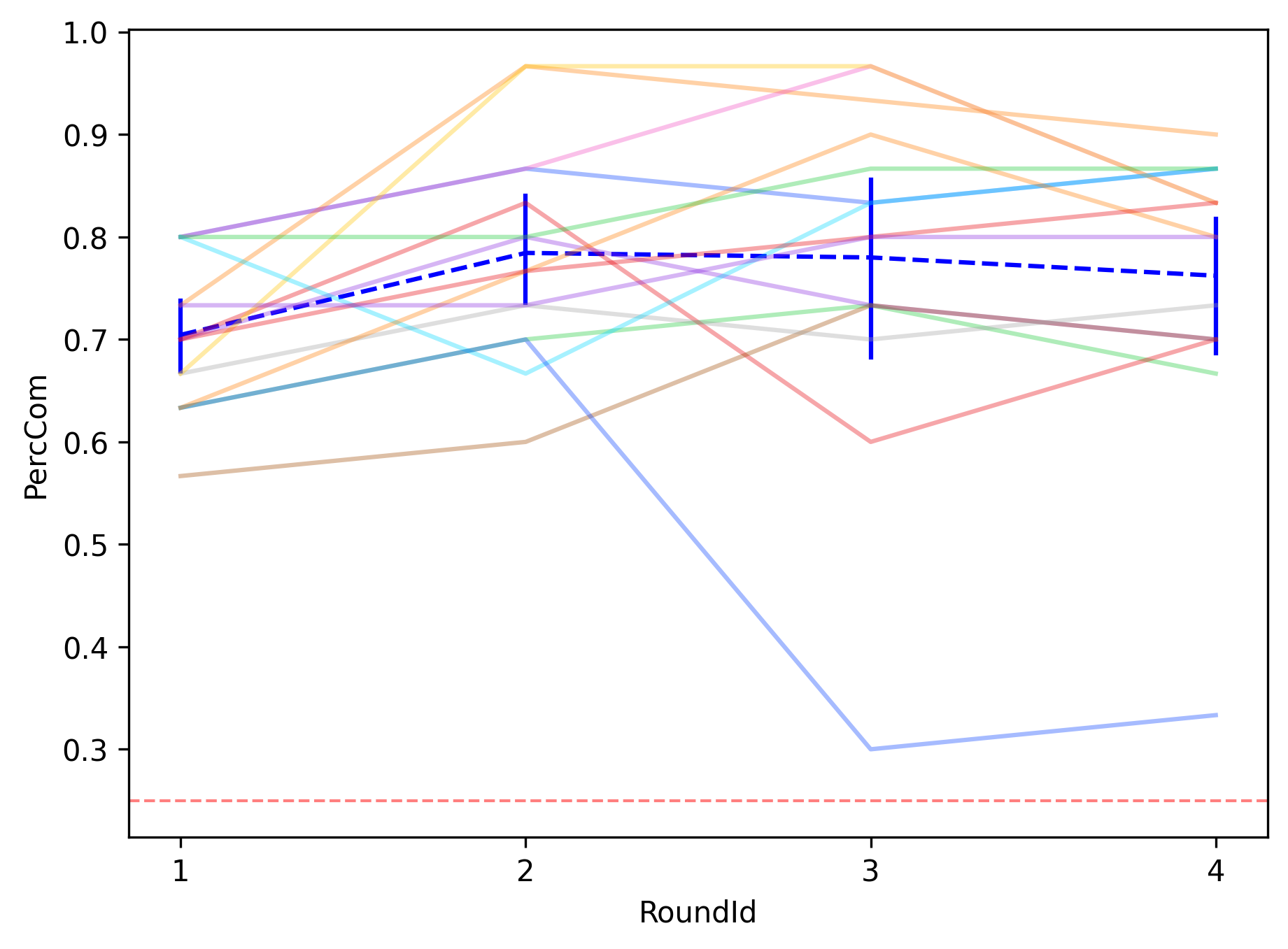}
    \caption{The communicative success (\textit{PercCom}) over the communication rounds. Each line indicates a simulation, the dashed blue line is the average with bars indicating the 95\% confidence interval. The dashed red line indicates chance performance.}
    \label{fig:percCom}
\end{figure}

\section{Additional results iterated learning}
\label{sec:iterated-learning}

\begin{figure}[ht]
    \centering
    \includegraphics[width=1\linewidth]{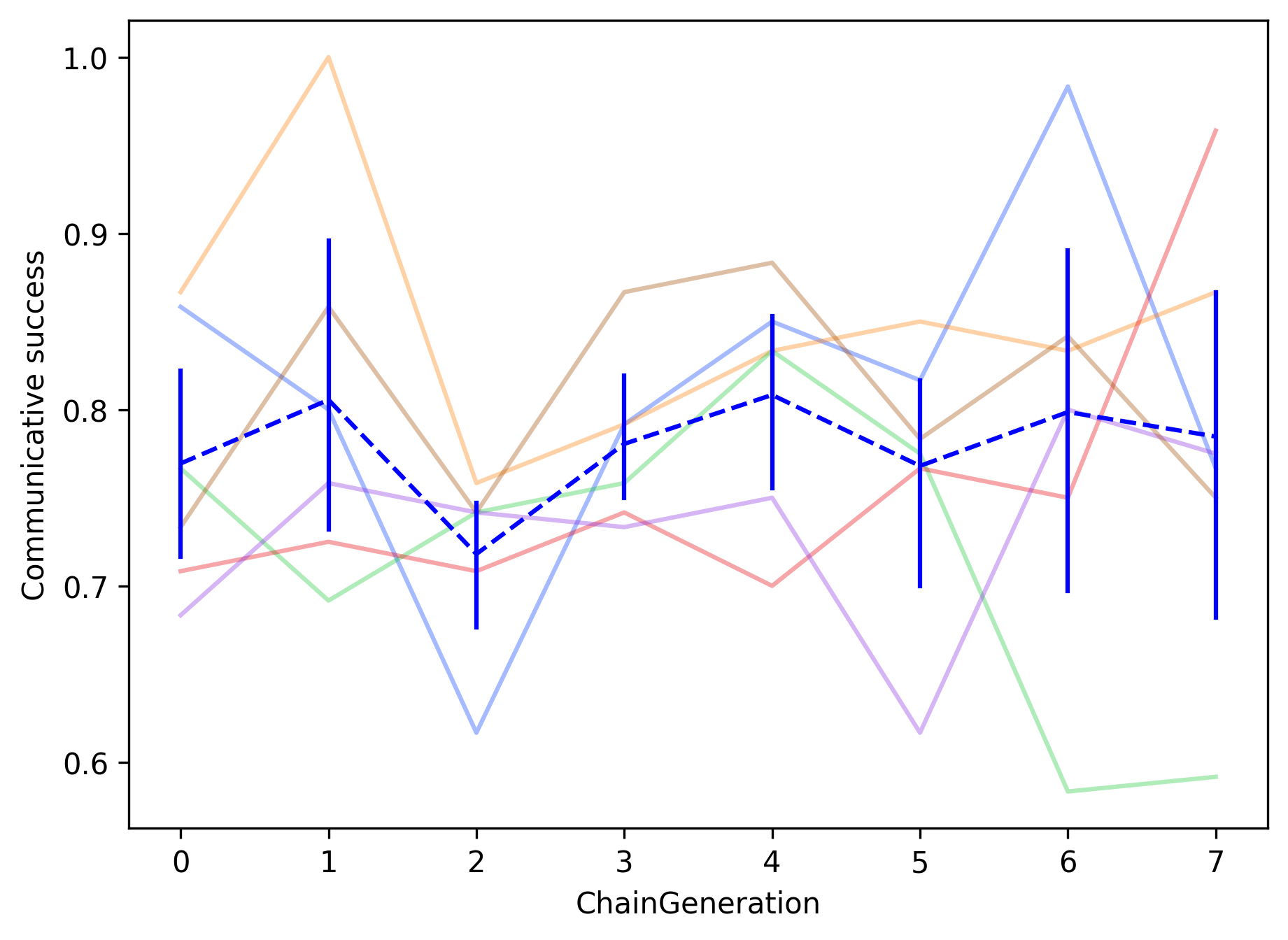}
    \caption{The average communicative success across rounds for each generation. Each line indicates a chain, and the dashed blue line is the average with bars indicating the 95\% confidence interval. See Table \ref{tab:ttests}.}
    \label{fig:chain-PercCom}
\end{figure}

\begin{figure}[ht]
    \centering
    \includegraphics[width=1\linewidth]{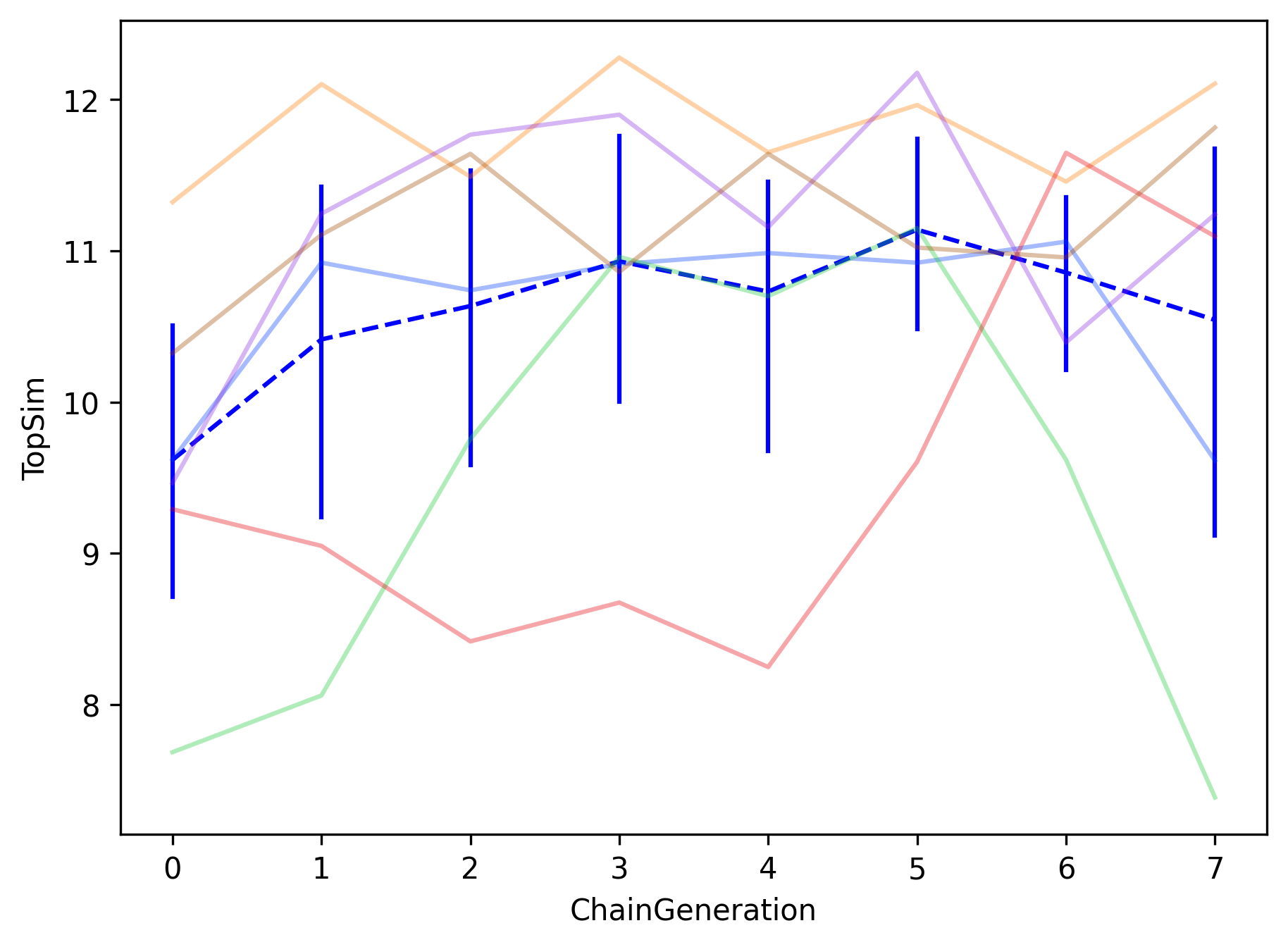}
    \caption{The evolution of \textit{TopSim} on the words produced in the testing block. Each line indicates a chain, and the dashed blue line is the average with bars indicating the 95\% confidence interval. See Table \ref{tab:ttests} for the descriptives of \textit{TopSim} and \textit{Ngram}.}
    \label{fig:Structure-Chain}
\end{figure}

\newpage
\section{Prompts}
\label{sec:prompts}
Our agents act based on prompts and system instructions. These are designed to be maximally close to the classical experimental setup and formatted similar to \citet{Galke2023WhatMakes}. The completion prompt \ref{fig:full-cmpl-prompt-labelling} is used for labelling and guessing. For the guessing task, we prefill the word and pick the signal with the highest probability. See the full prompts for labelling and guessing  (Prompt \ref{fig:full-cmpl-prompt-labelling}), speaking (Prompt \ref{fig:full-discr-prompt-speaker}), discrimination (Prompt \ref{fig:full-discr-prompt-listener}) below.

\begin{prompt*}[ht]
\small
\begin{lstlisting}[mathescape=true,escapeinside={*@}{@*}]
<|begin_of_text|><|start_header_id|>system<|end_header_id|> You are a language learner who has to learn an artificial language with words and their corresponding features. Your task is to complete the vocabulary by generating a word that describes the last item. Only respond with the word.<|eot_id|><|start_header_id|>user<|end_header_id|> 

{'shape':2,'colour':'orange','amount':1,'word':'giniwite'}
{'shape':3,'colour':'green','amount':1,'word':'ginisu'}
{'shape':1,'colour':'orange','amount':2,'word':'pinisugi'}
{'shape':3,'colour':'green','amount':3,'word':'sutepi'}
{'shape':2,'colour':'orange','amount':2,'word':'winisu'}
{'shape':3,'colour':'orange','amount':1,'word':'niwi'}
{'shape':1,'colour':'blue','amount':2,'word':'sutuwite'}
{'shape':1,'colour':'blue','amount':3,'word':'tupitene'}
{'shape':3,'colour':'blue','amount':1,'word':'wipinepi'}
{'shape':2,'colour':'orange','amount':3,'word':'gigi'}
{'shape':1,'colour':'green','amount':2,'word':'nite'}
{'shape':3,'colour':'blue','amount':3,'word':'wite'}
{'shape':1,'colour':'green','amount':3,'word':'sune'}
{'shape':2,'colour':'blue','amount':2,'word':'ninene'}
{'shape':2,'colour':'green','amount':1,'word':'tusetetu'}
{'shape':1,'colour':'green','amount':3,'word':'<|eot_id|><|start_header_id|>assistant<|end_header_id|>
[COMPLETION OR PREFFILED]
\end{lstlisting}
\caption{Completion Prompt used for labelling and guessing.}\label{fig:full-cmpl-prompt-labelling}
\end{prompt*}

\begin{prompt*}[ht]
\small
\begin{lstlisting}[mathescape=true,escapeinside={*@}{@*}]
<|begin_of_text|><|start_header_id|>system<|end_header_id|> You are a language learner who has to learn an artificial language with words and their corresponding features. Your task is to generate a word such that your communication partner can guess the correct meaning of the word. Communicative success is important. Only respond with the word.<|eot_id|><|start_header_id|>user<|end_header_id|>

{'shape':1,'colour':'green','amount':3,'word':'sutupitite','communicativeSuccess':1}
{'shape':2,'colour':'orange','amount':2,'word':'ginupepi','communicativeSuccess':1}
{'shape':1,'colour':'orange','amount':2,'word':'sutupepi','communicativeSuccess':1}
{'shape':1,'colour':'green','amount':2,'word':'sutupepi','communicativeSuccess':0}
{'shape':2,'colour':'orange','amount':1,'word':'ginisu','communicativeSuccess':1}
{'shape':2,*@\hspace{-1pt}@*'colour':'orange',*@\hspace{-1pt}@*'amount':3,*@\hspace{-1pt}@*'word':'ginupitite',*@\hspace{-1pt}@*'communicativeSuccess':1}
{'shape':3,'colour':'green','amount':1,'word':'wipisu','communicativeSuccess':0}
{'shape':2,'colour':'green','amount':1,'word':'ginisu','communicativeSuccess':1}
{'shape':1,'colour':'blue','amount':2,'word':'sunupepi','communicativeSuccess':1}
{'shape':3,'colour':'green','amount':3,'word':'wipipitite','communicativeSuccess':1}
{'shape':3,'colour':'orange','amount':1,'word':'wipisu','communicativeSuccess':0}
{'shape':1,'colour':'blue','amount':3,'word':'sunupitite','communicativeSuccess':1}
{'shape':3,'colour':'blue','amount':3,'word':'wipipitite','communicativeSuccess':1}
{'shape':3,'colour':'blue','amount':1,'word':'wipisu','communicativeSuccess':1}
{'shape':2,'colour':'blue','amount':2,'word':'<|eot_id|><|start_header_id|>assistant<|end_header_id|>
[COMPLETION]
\end{lstlisting}
\caption{Speaking Prompt during communication.}\label{fig:full-discr-prompt-speaker}
\end{prompt*}

% https://tex.stackexchange.com/questions/28556/how-to-place-a-float-at-the-top-of-a-floats-only-page?answertab=active#tab-top
% \makeatletter
% \setlength{\@fptop}{0pt}
% \setlength{\@fpbot}{0pt plus 1fil}
% \makeatother
% \bigskip

\begin{prompt*}[!t]
\small
\begin{lstlisting}[mathescape=true,escapeinside={*@}{@*}]
<|begin_of_text|><|start_header_id|>system<|end_header_id|> You are a language learner who has to learn an artificial language with words and their corresponding features. Your task is to complete the vocabulary by interpreting the intended meaning of the word generated by your communication partner. Communicative success is important. Only respond with the complete last item.<|eot_id|><|start_header_id|>user<|end_header_id|> 

{'word':'wipipitite','shape':3,'colour':'blue','amount':3,'communicativeSuccess':1}
{'word':'wipisu','shape':3,'colour':'orange','amount':1,'communicativeSuccess':0}
{'word':'wipisu','shape':3,'colour':'green','amount':1,'communicativeSuccess':0}
{'word':'sutupepi','shape':1,'colour':'orange','amount':2,'communicativeSuccess':1}
{'word':'ginupepi','shape':2,'colour':'orange','amount':2,'communicativeSuccess':1}
{'word':'sutupitite','shape':1,'colour':'green','amount':3,'communicativeSuccess':1}
{'word':'wipipitite','shape':3,'colour':'green','amount':3,'communicativeSuccess':1}
{'word':'wipisu','shape':3,'colour':'blue','amount':1,'communicativeSuccess':1}
{'word':'ginisu','shape':2,'colour':'green','amount':1,'communicativeSuccess':1}
{'word':'ginisu','shape':2,'colour':'orange','amount':1,'communicativeSuccess':1}
{'word':'sunupepi','shape':1,'colour':'blue','amount':2,'communicativeSuccess':1}
{'word':'sutupepi','shape':1,'colour':'green','amount':2,'communicativeSuccess':0}
{'word':'sunupitite','shape':1,'colour':'blue','amount':3,'communicativeSuccess':1}
{'word':'ginupitite',*@\hspace{-1pt}@*'shape':2,*@\hspace{-1pt}@*'colour':'orange',*@\hspace{-1pt}@*'amount':3,*@\hspace{-1pt}@*'communicativeSuccess':1}
{'word':'ginupepi','shape':'<|eot_id|><|start_header_id|>assistant<|end_header_id|>
[PREFILLED WITH DISTRACTOR ATTRIBUTES]
\end{lstlisting}
\caption{Guessing Prompt during communication.}\label{fig:full-discr-prompt-listener}
\end{prompt*}

\begin{prompt*}[!b]
\begin{lstlisting}[escapeinside={*@}{@*}]
 






















     
     
      *@\!@*                     
\end{lstlisting}
\end{prompt*}
          
\end{document}